\documentclass[conference]{IEEEtran}
\IEEEoverridecommandlockouts

\usepackage{cite}
\usepackage{amsmath,amssymb,amsfonts}
\usepackage{algorithmic}
\usepackage{graphicx}
\usepackage{textcomp}
\usepackage{xcolor}

\usepackage{multirow}

\def\BibTeX{{\rm B\kern-.05em{\sc i\kern-.025em b}\kern-.08em
    T\kern-.1667em\lower.7ex\hbox{E}\kern-.125emX}}
\begin{document}

\title{Combined CNN and ViT features off-the-shelf: \\ Another astounding baseline for recognition.
}

\author{\IEEEauthorblockN{Fernando Alonso-Fernandez, Kevin Hernandez-Diaz, Prayag Tiwari, Josef Bigun}
\IEEEauthorblockA{Halmstad University. Box 823. SE 301-18 Halmstad, Sweden\\
Emails:  feralo@hh.se, kevin.hernandez-diaz@hh.se, prayag.tiwari@hh.se, josef.bigun@hh.se}}

\maketitle

\begin{abstract}
We apply pre-trained architectures, originally developed for the ImageNet Large Scale Visual Recognition Challenge, for periocular recognition. 
These architectures have demonstrated significant success in various computer vision tasks beyond the ones for which they were designed.
This work builds on our previous study using off-the-shelf Convolutional Neural Network (CNN) and extends it to include the more recently proposed Vision Transformers (ViT).
Despite being trained for generic object classification, middle-layer features from CNNs and ViTs are a suitable way to recognize individuals based on periocular images.
We also demonstrate that CNNs and ViTs are highly complementary since their combination results in boosted accuracy.
In addition, we show that a small portion of these pre-trained models can achieve good accuracy, resulting in thinner models with fewer parameters, suitable for resource-limited environments such as mobiles.
This efficiency improves if traditional handcrafted features are added as well. 
\end{abstract}


\section{INTRODUCTION \& RELATED WORK}

The periocular region, 
the area
around the eye, is 
one of the most distinctive parts of the face \cite{Alonso24computers_periSOA}.
It offers flexibility and minimal acquisition constraints, requiring little user collaboration, making it ideal for non-cooperative biometrics.
This region represent a balance between using the iris (unavailable at long distances due to low resolution, e.g. surveillance) and the face (which may be partially occluded, either accidentally, such as in selfies \cite{[Rattani17soaOcularVIS]} or masks \cite{sharma23cviu_periocular_masks_survey}, or intentionally, such as criminals concealing their face). 
All these properties make the periocular region an excellent biometric modality on its own, having been used for other tasks too beyond identity recognition, e.g. soft-biometrics or expression estimation \cite{Alonso24computers_periSOA}.

Convolutional Neural Networks (CNNs), and more recently Vision Transformers (ViTs), have become popular tools in vision tasks, including biometrics \cite{minaee23AIR_DL_biometrics_survey,Khan22ACMCS_ViTsurvey}. 
A common limitation is the need for substantial and varied training data, \cite{Luo18icsai_data_affects_CNN}, being transfer learning (TL) one of the most common approaches to tackle data scarcity.
In TL, a network trained on a more complex task is adapted for a new task, leveraging the extensive training data of the original domain and feature extraction power of pre-trained CNNs.
One-shot learning, an extreme form of TL, uses the network without modification \cite{Hernandez23access_oneshot}, utilizing embedding vectors as feature vectors (usually from the last layer, but not always) to compare images via distance or similarity metrics.
A popular choice is the networks from the ImageNet Large Scale Visual Recognition Challenge\footnote{https://www.image-net.org/}, trained on over a million images to classify images into 1000 common object categories, producing generic feature descriptors shown to be powerful for many vision tasks \cite{[Razavian14]}.

Inspired by previous works in iris and periocular \cite{[Nguyen18],[Hernandez18],[Hernandez19],[Alonso22inffus],Hernandez23access_oneshot}, we leverage these architectures for biometrics, more particularly periocular recognition. 
Previously, CNNs were employed for this task.
Building on prior work \cite{[Hernandez18]}, we now include ViTs alongside CNNs, adding ViTs to the evidence supporting the use of such generic descriptors.
We compare three CNNs and three ViTs with varying depths and parameters to analyze the impact of network complexity. We also explore different methods to normalize layer representations.

\begin{figure}[b]
\centering
\includegraphics[width=0.4\textwidth]{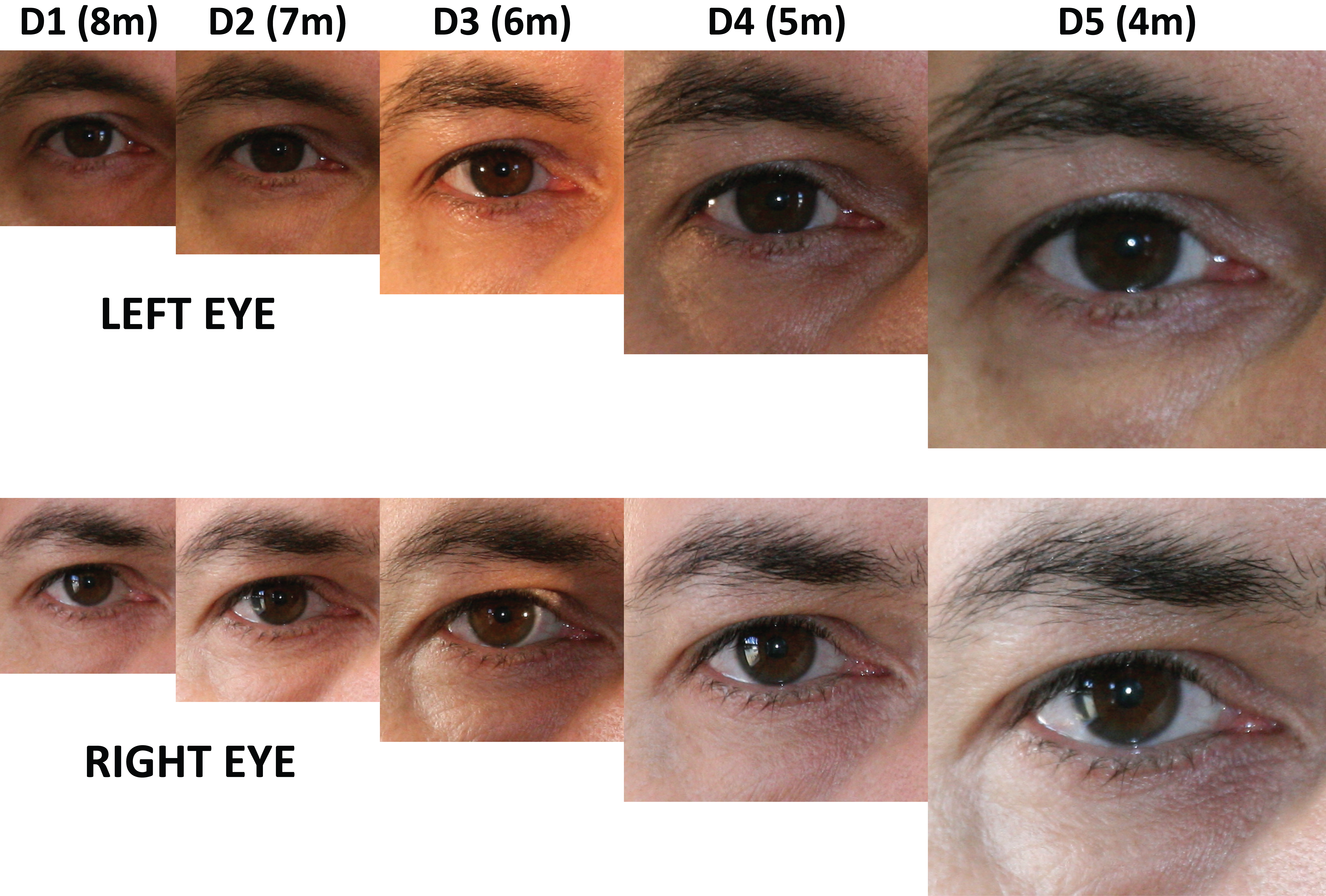}
\caption{Images from one user of the UBIPr database (relative scale difference between images is shown). Note that images of the right eye are flipped horizontally to match the same orientation as the left image (Section~\ref{sect:db_protocol}).} \label{fig:db-samples}
\end{figure}

Moreover, we show that combining CNN and ViT descriptors is even more powerful, showing that they are highly complementary.
To close the circle of complementarity, we include traditional Local Binary Patterns (LBP) \cite{[Ojala02]}, Histogram of Oriented Gradients (HOG) \cite{[Dalal05]} and Scale-Invariant Feature Transform (SIFT) key-points \cite{[Lowe04]}, showing that these hand-crafted descriptors can also benefit the recognition mission.

In addition, we eliminate the advantage observed in previous works, including ours, where left and right eye images were compared directly. Instead, we flip images of one eye horizontally (Figure~\ref{fig:db-samples}), so the orientations and shapes of the overall eye structure, tear duct positions, eyelids and eyelashes shape, etc. are aligned.
This is a more difficult but more realistic task since, in uncontrolled scenarios like selfies or social media, face pictures may appear flipped for different reasons.

\begin{figure}[t]
\centering
\includegraphics[width=0.45\textwidth]{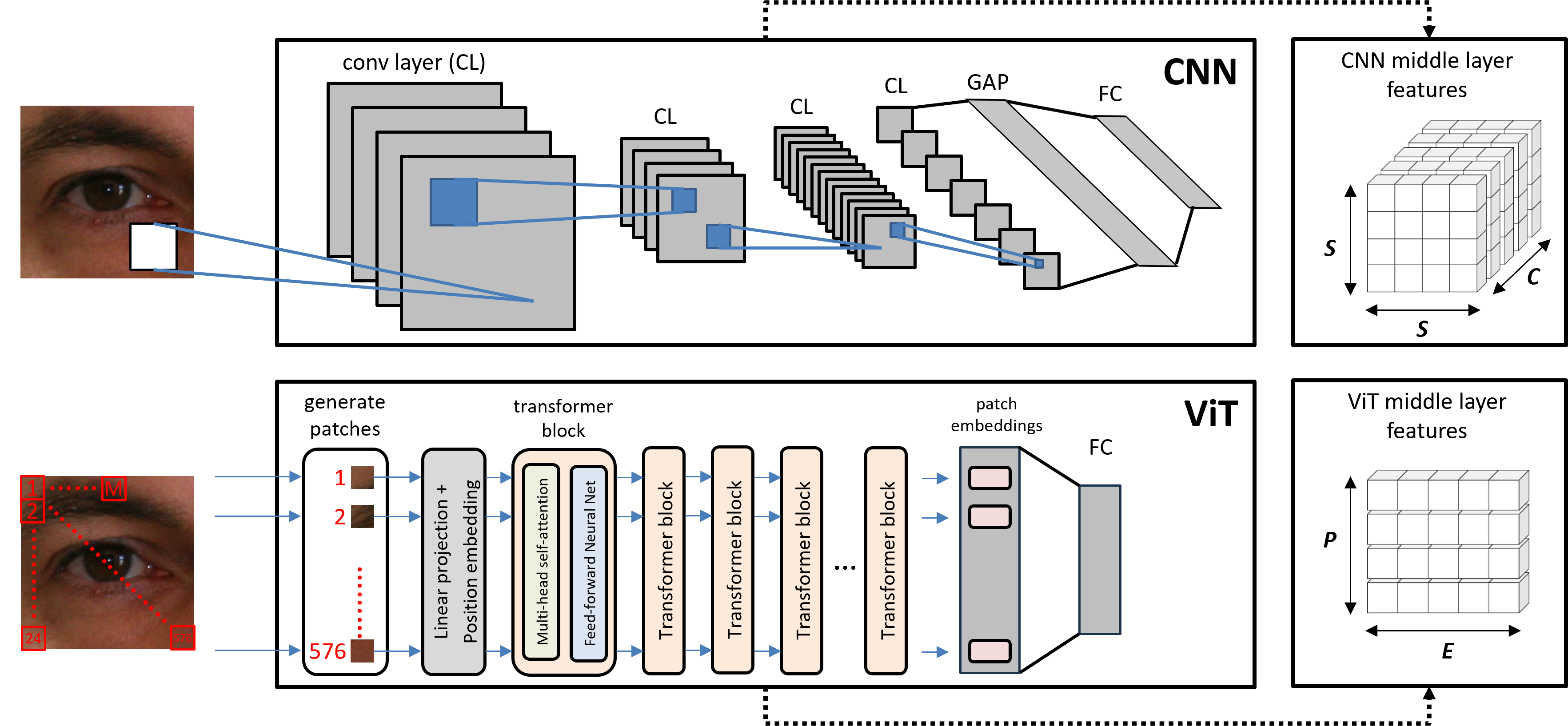}
\caption{Extraction of periocular features from different CNN/ViT layers.} \label{fig:cnns}
\end{figure}

\begin{table}[t]
\small
    \caption{CNNs and ViTs evaluated in this paper.}
    \centering
    \begin{tabular}{|c|c|c|c|c|c|}

    \hline
    \textbf{} &  \textbf{Conv} &   \textbf{Size} & \textbf{Para-} & \textbf{Spatial} & \textbf{Activation}  \\

    \textbf{CNN}  & \textbf{Layers}  & \textbf{MB} & \textbf{meters} & \textbf{Size $S$} & \textbf{Channels $C$}  \\
    \hline \hline


R18 & 18 &  44  & 11.7M & 112,56 & 64,128,256,512  \\ \cline{1-4} \cline{6-6}

R50 & 50 &  96 & 25.6M & 28,14 & 64,128,256  \\  \cline{1-4} 

R101 & 101 &  167  & 44.6M & 7 & 512,1024,2048  \\ \hline









\multicolumn{6}{c}{} \\

\hline
\textbf{} &  \textbf{Transf} &   \textbf{Size} & \textbf{Para-} & \textbf{Patches} & \textbf{Embedding}  \\

    \textbf{ViT}  & \textbf{Blocks}  & \textbf{MB} & \textbf{meters} & \textbf{$P$} & \textbf{Size $E$}  \\
    \hline \hline

tiny & 12 &  20.6  & 5.7M & 577 & 192, 768    \\ \hline

small & 12 &  78.8  & 22.1M & 577 & 384, 1536   \\ \hline

base & 12 &  308  & 86.8M & 577 & 768, 3072  \\ \hline

    \end{tabular}
    \label{tab:networks_used}
\end{table}

\begin{figure*}[t]
\centering
\includegraphics[width=0.95\textwidth]{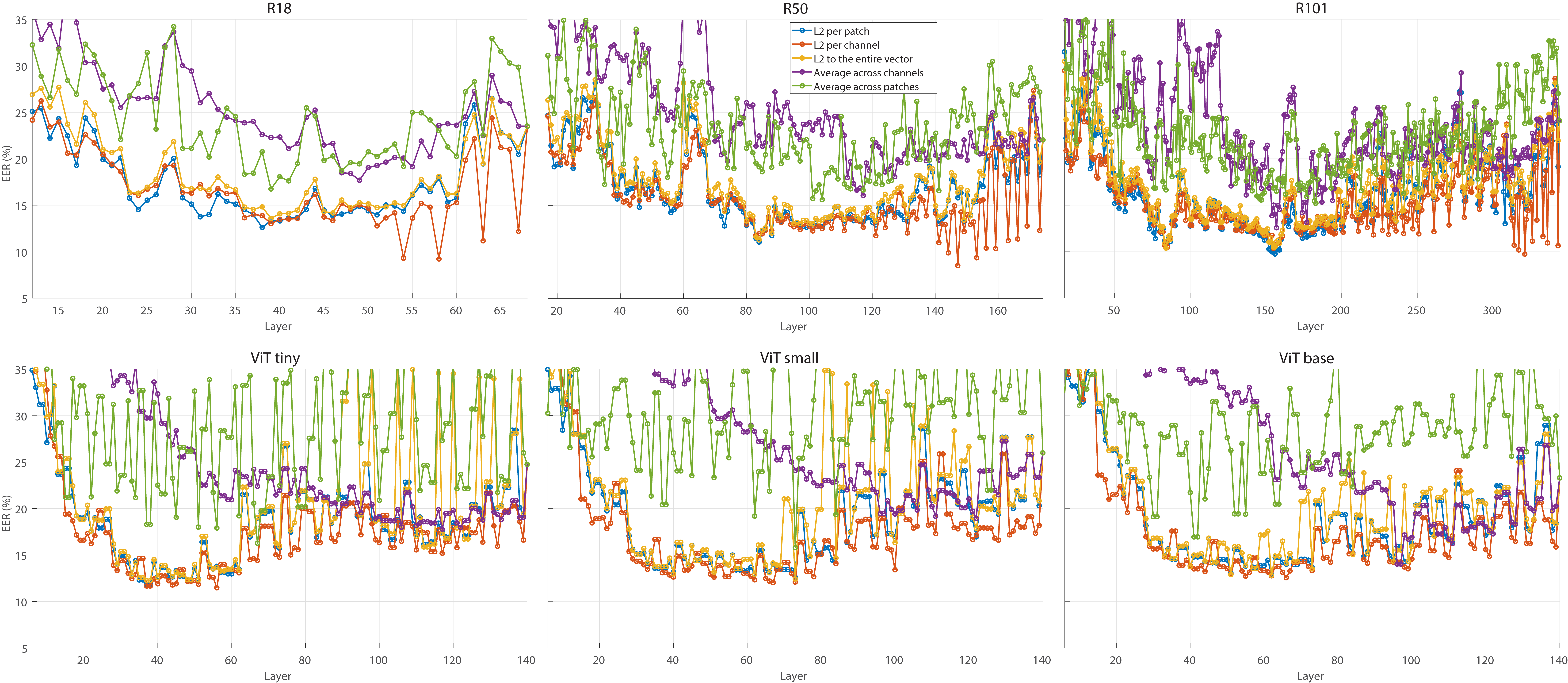}
\caption{Verification results (EER) of different CNN and ViT layers with the various vector normalization techniques employed.} \label{fig:resuls-individual}
\end{figure*}

\section{METHODOLOGY}

\subsection{CNN and ViT features}
\label{subsect:networks}

%
%

We employ three widely used CNNs and three ViTs (Table~\ref{tab:networks_used}), available in Matlab r2023b.
They are chosen for their varying parameters and depths.
We use three CNN ResNet variants (R18, R50 and R101) \cite{[He16]}, differing in convolutional layers and activation channels per layer. 
ResNet models are popular in biometrics and facial modalities \cite{Deng19CVPR_ArcFace}.
%
%
%
%
%
%
For ViTs, we use three variants (tiny, small, and base) \cite{dosovitskiy21iclr_ViT_transformers}, which share the same number of transformer blocks but differ in the number of channels (embedding size) across each block.

In using these networks, periocular images are fed into the feature extraction pipeline of each pre-trained model.
We use vectors from intermediate layers as feature descriptors (Figure~\ref{fig:cnns}).
In CNNs, it is a 3D tensor of size $S \times S \times C$, consisting of $C$ activations (channels) of spatial size $S \times S$. We extract features from all layers (convolutional, pooling, batch norm, ReLU, addition, fully connected, etc.).
In ViTs, the middle vector is a 2D tensor of size $P \times E$, being $P$ the number of patches 
and $E$ the patch embedding size.
%
%
Table~\ref{tab:networks_used} details the parameters. Larger models like R101 and base ViT use more activations $C$ and larger embeddings $E$.
%
%
%
The last layers, typically global average pooling or fully connected layers, are exceptions, with $S$=1 or $P$=1.

The extracted CNN and ViT vectors are compared pairwise via cosine similarity, a common method in biometrics \cite{Deng19CVPR_ArcFace}. Cosine similarity involves L2-normalizing the vectors to focus on their angle. In doing so, we evaluated several possibilities:

\begin{enumerate}

    \item \textit{L2 normalization per patch}. Each patch of the feature vector is normalized separately, considering all channels. Cosine similarity is computed per patch, then averaged for a single score. For CNNs, the feature vector is sliced into $S \times S$ sub-vectors of size $1 \times 1 \times C$. For ViTs, each of the $P$ rows of size $1 \times E$ is normalized separately.

    \item \textit{L2 normalization per channel}. This normalization integrates information across all patches for a given channel. For CNNs, the feature vector is sliced into $C$ subvectors of size $S \times S$. For ViTs, each of the $E$ columns of the feature vector of size $P \times 1$ is normalized separately. 

    \item \textit{L2 normalization of the entire vector} by its L2-norm, maintaining relative magnitude differences without per-patch or per-channel normalization.

    \item \textit{Average across channels}. The feature vector is averaged across channels and then L2-normalized, mimicking Global Average Pooling usually done in the last layer of CNNs before the fully connected. The CNN vector size is $1 \times C$, and the ViT vector is $1 \times E$.

    \item \textit{Average across patches}. Similar to the previous method but averaging across patches. The CNN vector size is $S \times S \times 1$, and the ViT vector is $P \times 1$.    

\end{enumerate}

\subsection{Traditional features}
\label{subset:traditional-features}

We also use three traditional hand-crafted feature methods, employed as baseline in many periocular studies \cite{[Alonso16]}: LBP \cite{[Ojala02]},
HOG \cite{[Dalal05]},
and SIFT \cite{[Lowe04]}.
In HOG and LBP, images are divided into 8$\times$8 non-overlapping 
regions, and features are extracted from each region, forming per-block histograms which are concatenated to form a feature vector of the whole image. 
Vector comparison 
is done via distance metrics ($\chi^2$ in this work \cite{[Hernandez18]}).
LBP and HOG are extracted using Matlab functions \texttt{extractLBPFeatures} (with the rotation invariance flag \textit{Upright} set to true) and \texttt{extractHOGFeatures} (with the \textit{UseSignedOrientation} flag set to true).
On the other hand, SIFT extracts key-points (with dimension 128 per key-point) from the entire image, with recognition based on the number of paired key-points, normalized by the smallest number of detected key-points in the images.
We use a free implementation\footnote{http://vision.ucla.edu/ vedaldi/code/sift/assets/sift/index.html}, adapted to filter out spurious pairings by constraining the angle and distance of paired key-points \cite{[Alonso09]}.

\subsection{Database and protocol}
\label{sect:db_protocol}

We use the UBIPr periocular database \cite{[Padole12]}, which contains images from a CANON EOS 5D camera taken with varying distance (4-8m) and resolution (501$\times$401 at 8m to 1001$\times$801 pixels at 4m). 
%
We select 1,718 frontal-view images from 86 individuals (users with two sessions), with 2 images per eye and per distance, totaling 172$\times$2=344 images per distance. 
%
%
Images are manually annotated for pupil and sclera circle parameters, resized to match the average sclera radius $R_s$ of its distance group, and aligned by extracting a $7.6 R_s
\times 7.6 R_s$ square around the sclera center. 
We use the sclera for normalization since it is not affected by dilation compared to the the pupil.
They are then resized to match the input size of the CNN or ViT. Unlike some methods \cite{[Park11]}, and to provide a more realistic application, we do not mask the iris. 
Iris segmentation can be challenging in real-world conditions due to low resolution, motion blur, reflections of low pigmentation in visible images, which is precisely where the periocular image and its less demanding acquisition 
stand out \cite{Alonso24computers_periSOA}.
%
An example of normalized images is given in Figure
\ref{fig:db-samples}.
%

%
%
%
%
%
%
%
%

We perform verification experiments, with each eye considered a different user.
As mentioned earlier, to eliminate the advantage of a different orientation between left and right eyes, we flip right images horizontally.
%
For genuine trials, we compare all images of the same user, avoiding symmetric comparisons, resulting in 7,722 user scores.
For impostor trials, we use the first image of a user as enrolment and compare it with the second image of the
remaining users, producing 172$\times$171=29,412 scores.
%
We also conduct fusion using linear logistic regression to combine scores from multiple comparators. Given $N$ comparators with scores ($s_{1j}, s_{2j}, ... s_{Nj}$) for an input trial $j$, a linear fusion 
is $f_j = a_0 + a_1 \cdot
s_{1j} + a_2 \cdot s_{2j} + ... + a_N \cdot s_{Nj}$. The weights
$a_{0}, a_{1}, ... a_{N}$ are trained via logistic regression
\cite{[Alonso08]}. 
This trained fusion has been shown to outperform simple fusion rules like the mean or sum \cite{[Alonso08]}.
Nonetheless, this is a weighted sum, though the coefficients are optimized by a specific rule \cite{[Bigun97]}.

\begin{figure}[t]
\centering
\includegraphics[width=0.45\textwidth]{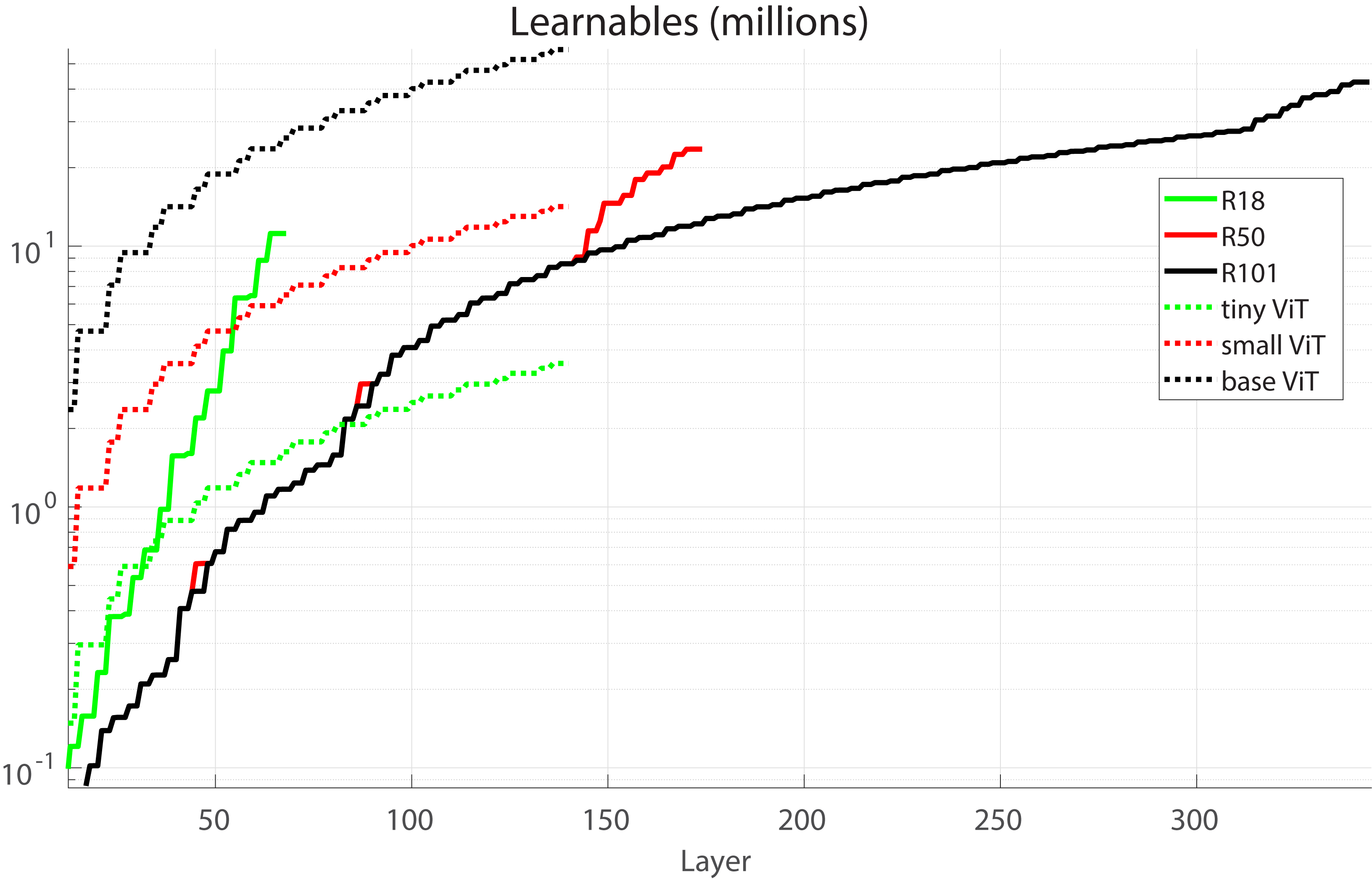}
\caption{Learnables of the CNNs and ViTs up to a certain layer.} \label{fig:learnables}
\end{figure}

\begin{figure*}[htb]
\centering
\includegraphics[width=0.95\textwidth]{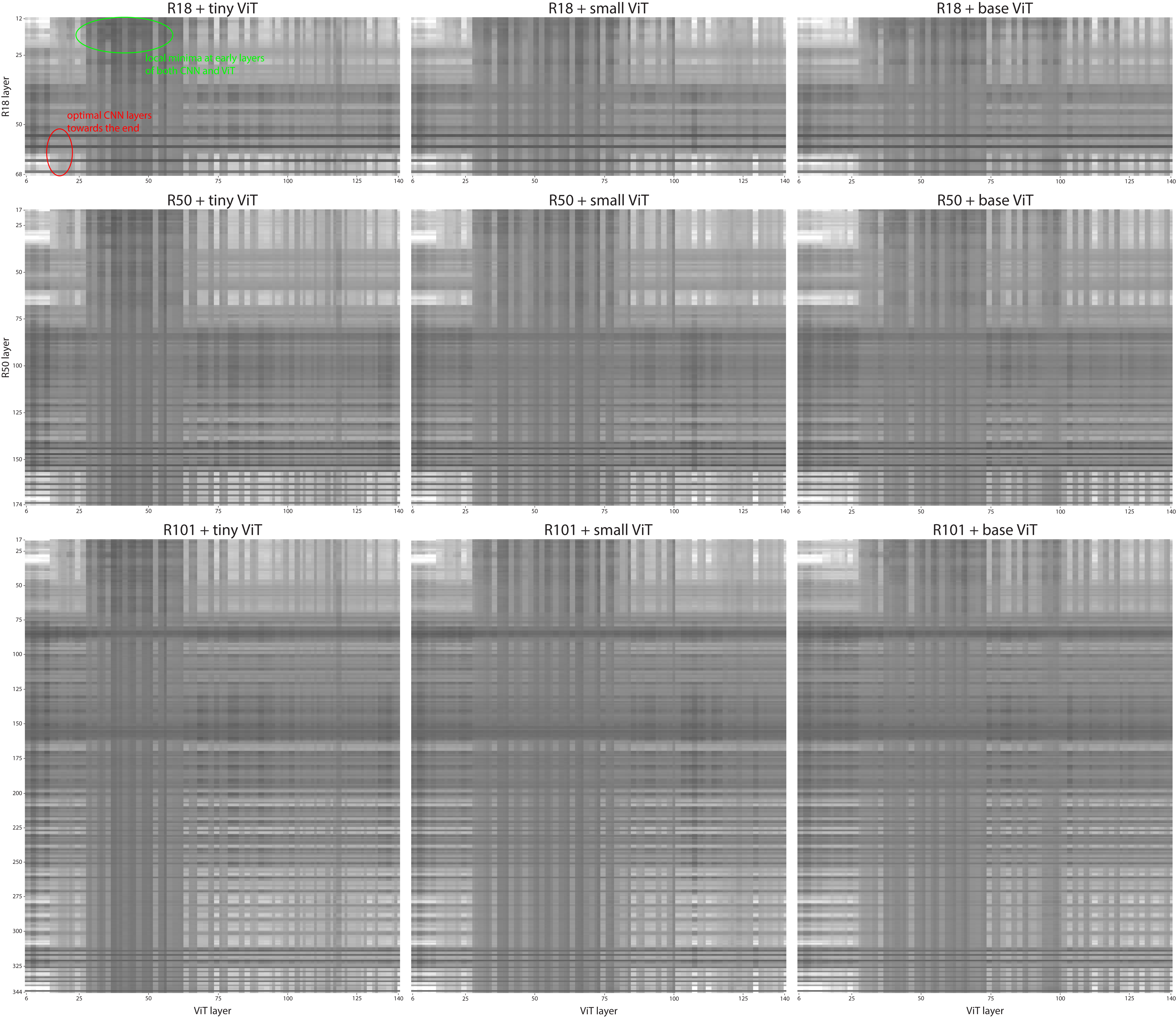}
\caption{Fusion results of CNN with ViT layers using L2 normalization per channel (optimal case of Figure~\ref{fig:resuls-individual}). Grey values of the images are scaled between 0\% (black) and 25\% EER (white).} \label{fig:resuls-fusion}
\end{figure*}


\begin{figure*}[t]
\centering
\includegraphics[width=0.95\textwidth]{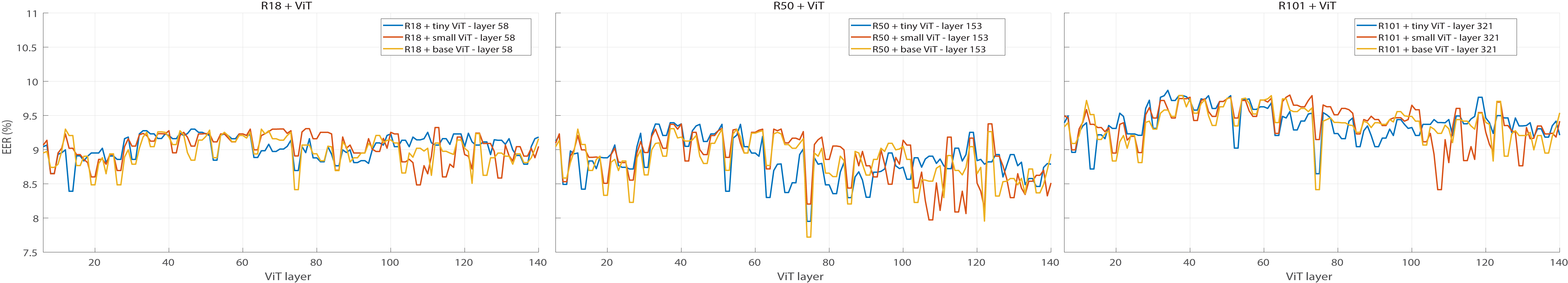}
\caption{Fusion results of selected CNN layers (best individual ones of Table~\ref{tab:eer-individual}) against all ViT layers.} \label{fig:resuls-fusion-selection}
\end{figure*}

\begin{table}[]
\centering
\caption{Verification results (EER) of the traditional systems (left), of the best layer of each CNN (center), and of the best layer of each ViT (right). The best performance given the CNNs and ViTs in our experiments is marked in bold. Parameters (in millions) and layer involved are also given. (*) According to \cite{[Hernandez18]}.}
\label{tab:eer-individual}
\resizebox{\columnwidth}{!}{%
\begin{tabular}{p{0.4cm}p{0.4cm}cp{0.5cm}p{0.25cm}p{0.3cm}p{0.6cm}cp{0.3cm}p{0.3cm}p{0.3cm}p{0.6cm}}
\multicolumn{2}{c}{traditional} &  & \multicolumn{4}{c}{CNNs} &  & \multicolumn{4}{c}{ViTs} \\ \cline{1-2} \cline{4-7} \cline{9-12}

\textbf{} & \textbf{EER} &  & \textbf{} & \textbf{EER} & \textbf{layer} & \textbf{param.} &  & \textbf{} & \textbf{EER} & \textbf{layer} & \textbf{param.} \\

LBP & 19.37 &  & R18 & 9.23 & 58 & 6.33 &  & tiny & \textbf{11.48} & 56 & 1.33 \\

HOG & 16.61 &  & R50 & \textbf{8.53} & 147 & 11.45 &  & small & 12.01 & 67 & 6.50 \\

SIFT & 17.26 &  & R101 & 9.74 & 321 & 31.52 &  & base & 12.56 & 66 & 23.61 \\ \cline{4-7} \cline{9-12}

all & 10.58 &  & R50* & 6.4 & 73 & - &  &  &  &  & \\ \cline{1-2}

all* & 9.1 &  & R101*  & 5.6 & 170 & - &  &  &  &  & \\ \cline{4-7}

all* & 16 &  &  &  &  &  &  &  &  &  & \\ \cline{1-2}

\end{tabular}
}
\end{table}

\section{EXPERIMENTAL RESULTS}


We first examine the representation capability of each CNN and ViT layer by reporting verification accuracy using features from each layer (Figure~\ref{fig:resuls-individual}).
Averaging the feature vector across channels or patches (purple/green curves) reduces recognition accuracy, suggesting that such channel or patch integration method is ineffective. 
Among the other three normalization techniques, L2 normalization per channel (red curves) yields the best accuracy, especially with CNNs. 
On the other hand, L2-normalizing the entire vector (yellow curves) generally performs the worst, particularly with ViTs. 
%
There are some layers where the three normalization techniques do not differ much, or where L2 normalization per patch (blue curves) is even better. 
However, since L2 normalization per channel 
gives the best absolute performance per network, we will move forward only with this normalization in the remainder of the paper.
The accuracy for each network at the best-performing layer with this normalisation is reported in Table~\ref{tab:eer-individual}.

With CNNs, the best performance happens towards the end of the network (layers 58 in R18, 147 in R50, 321 in R101). 
An exception is R101, with a local minima at 84 and 156 nearly matching the absolute minimum at 321. 
This indicates that while R18 and R50 require most layers for optimal performance, R101 does not need to go as deep.
Interestingly, R50 and R101 have a minimum around similar depths (147 vs. 156), and R50 also has a local minimum at 84. 
Further analysis of the networks' learnables (Figure~\ref{fig:learnables}) reveals that R50 and R101 have comparable parameters up to layer 150, which may explain that they share local minima at the same depths. 
%
%
%
At layer 147, R50 has 11.45M parameters, and at layer 156, R101 has 10.53M.
The learnables of the smaller R18 at layer 58 is also high (6.33M) since we are engaging the entire network, but this is 55\% or less than the other ResNet versions. 
Overall, R50 provides the best accuracy (EER=8.53\%) followed by R18 (9.23\%). 
It is also relevant that despite fewer parameters, R18 performance is only 0.7\% worse.
%

Contrarily to CNNs, the best performance of ViTs happens in the first half of the networks. 
The EER curves show a U-shape, decreasing to a minimum at layers 56-67 out of 140 (Table~\ref{tab:eer-individual}), with near-optimal performance already at layer 40 (less than 30\% of the layers).
This indicates that ViTs achieve sufficient abstraction earlier, and deeper layers degrade performance. 

Regarding learnables, 
at the optimal layer (Table~\ref{tab:eer-individual}), the tiny ViT has just 1.33M parameters, $\sim$20\% of the small and $\sim$6\% of the base ViT, yet their performance is similar, with the tiny ViT performing best.
%
%
This highlights that larger models are unnecessary for ViTs.
It also says about the impressive capabilities of ViTs, since the performance of the tiny ViT (11.48\% EER) is commendable compared to the R18 CNN (9.23\%) which has nearly 5$\times$ parameters.

Table~\ref{tab:eer-individual} also includes the results from our previous study with the same database \cite{[Hernandez18]}, where performance was better. 
However, the present study introduces some modifications, the most relevant being the mentioned alignment of left and right images, which is expected to make recognition more difficult.

We then study the complementarity between CNN and ViT features through fusion experiments involving all possible combinations of CNN and ViT layers, with results given in Figure~\ref{fig:resuls-fusion}. 
Table~\ref{tab:eer-fusion} (top) provides the EER of the best fusion cases for each CNN-ViT combination, along with results incorporating the traditional features from Section~\ref{subset:traditional-features}.

A first look at Table~\ref{tab:eer-fusion} (top) compared to Table~\ref{tab:eer-individual} reveals that the best fusion cases usually involve the optimal CNN layers or those nearby. They correspond to the oscillatory EER minima at the end of CNNs in Figure~\ref{fig:resuls-individual}, which translates to the horizontal dark bars in Figure~\ref{fig:resuls-fusion} (marked in red in the R18 + tiny ViT plot, but present in all combinations). 
Other horizontal dark bars align with local minima in the CNNs, 
such as layer 84 (R50) or layers 84 and 156 (R101). 
However, these dark bars span across all ViT layers, indicating that the ViT layer employed in the fusion is not as critical as the CNN layer.
This is further illustrated in Figure~\ref{fig:resuls-fusion-selection}, where the fusion of selected CNN layers with all ViT layers shows that the EER oscillates within a band of less than 1\%, regardless of the ViT layer.

Regarding EER values, Table~\ref{tab:eer-fusion} shows that in all cases, the fusion CNN + ViT improves performance (column `CNN+ ViT').
As observed earlier, R50 is the best CNN and tiny the best ViT (Table~\ref{tab:eer-individual}).
The best fusion combination is R50 CNN + base ViT (7.72\%), with R50 + tiny ViT as the runner-up (7.95\%). 
We also observed earlier that R18 is a bit behind R50 but with 60\% of the parameters. Also, here, the fusion of R18 with ViTs shows an EER only 0.6\% worse (8.32\%).

Another look at Figure~\ref{fig:resuls-fusion} reveals another dark region in the early layers of both CNN and ViTs (we mark an example in green). 
This suggests that another layer combination with good performance but fewer parameters might be possible. This would be useful e.g. in mobile biometrics \cite{alonsofernandez2024deepnetworkpruningcomparative} where computing and storage are limited.
By manually searching for local minima in these regions (results shown in Table~\ref{tab:eer-fusion}, bottom), we found that the best EER (R50+tiny) is 8.79\%, using only 1.03M parameters. 
This EER is approximately 1\% worse than the absolute minimum (with 16.38M parameters), but the required parameters are substantially fewer.

Finally, we incorporate the traditional features of Section~\ref{subset:traditional-features} (last column of Table~\ref{tab:eer-fusion}).
As it can be observed, the performance in all cases is further improved.
Notably (Table~\ref{tab:eer-individual}), the combined performance of these traditional features falls between the CNNs and the ViTs, so their role and contribution should not be dismissed, as seen in the fusion results. 
Additionally, the gap between the top and bottom parts of the table is reduced by adding these traditional systems, i.e. the overall minimums are 6.32\% and 6.65\%. 
This indicates that incorporating traditional features makes it easier to achieve models with fewer parameters without significantly sacrificing performance.


\begin{table}[]
\centering
\caption{Verification results (EER) of the fusion experiments. Values in bold correspond to the best EER of the column.}
\label{tab:eer-fusion}
\resizebox{0.95\columnwidth}{!}{%
\begin{tabular}{p{0.3cm}p{0.55cm}|p{0.3cm}p{0.3cm}|p{0.4cm}p{0.4cm}p{0.5cm}|p{0.4cm}p{0.4cm}p{0.5cm}p{0.6cm}}

\multicolumn{11}{c}{\textbf{selection by best EER of CNN+ViT}} \\ \hline

\multicolumn{2}{c|}{\textbf{system}} & \multicolumn{2}{c|}{\textbf{layer}} & \multicolumn{3}{c|}{\textbf{learnables}} & \multicolumn{4}{c}{\textbf{EER}} \\ \hline

\textbf{CNN} & \textbf{ViT} & \textbf{CNN} & \textbf{ViT} & \textbf{CNN} & \textbf{ViT} & \textbf{\begin{tabular}[c]{@{}c@{}}CNN\\ +ViT\end{tabular}} & \textbf{CNN} & \textbf{ViT} & \textbf{\begin{tabular}[c]{@{}c@{}}CNN\\ +ViT\end{tabular}} & \textbf{\begin{tabular}[c]{@{}c@{}}+tradi\\ tional\end{tabular}}  \\ \hline

\multirow{3}{*}{\textbf{R18}} & \textbf{tiny} & 54 & 13 & 3.97 & 0.15 & 4.11 & 9.32 & 25.60 & 8.32  & 6.55 \\

 & \textbf{small} & 58 & 107 & 6.33 & 10.63 & 16.96 & 9.23 & 25.07 & 8.48  & 6.83 \\
 
 & \textbf{base} & 54 & 74 & 3.97 & 28.33 & 32.30 & 9.32 & 17.88 & 8.33  & 6.78 \\  \hline
 
\multirow{3}{*}{\textbf{R50}} & \textbf{tiny} & 153 & 74 & 14.60 & 1.78 & 16.38 & 9.38 & 21.40 & 7.95  & 6.77 \\

 & \textbf{small} & 153 & 107 & 14.60 & 10.63 & 25.24 & 9.38 & 25.07 & 7.97  & 7.02 \\
 
 & \textbf{base} & 153 & 74 & 14.60 & 28.33 & 42.94 & 9.38 & 17.88 & \textbf{7.72}  & \textbf{6.32} \\  \hline
 
\multirow{3}{*}{\textbf{R101}} & \textbf{tiny} & 85 & 13 & 2.17 & 0.15 & 2.32 & 10.50 & 25.60 & 8.44  & 6.97 \\

 & \textbf{small} & 321 & 107 & 31.52 & 10.63 & 42.15 & 9.74 & 25.07 & 8.41  & 7.23 \\
 
 & \textbf{base} & 321 & 74 & 31.52 & 28.33 & 59.85 & 9.74 & 17.88 & 8.41  & 6.55 \\  \hline
 
\multicolumn{11}{l}{}  \\

\multicolumn{11}{c}{\textbf{selection by minimizing depth of CNN and ViT}} \\ \hline

\multicolumn{2}{c|}{\textbf{system}} & \multicolumn{2}{c|}{\textbf{layer}} & \multicolumn{3}{c|}{\textbf{learnables}} & \multicolumn{4}{c}{\textbf{EER}} \\ \hline

\textbf{CNN} & \textbf{ViT} & \textbf{CNN} & \textbf{ViT} & \textbf{CNN} & \textbf{ViT} & \textbf{\begin{tabular}[c]{@{}c@{}}CNN\\ +ViT\end{tabular}} & \textbf{CNN} & \textbf{ViT} & \textbf{\begin{tabular}[c]{@{}c@{}}CNN\\ +ViT\end{tabular}} & \textbf{\begin{tabular}[c]{@{}c@{}}+tradi\\ tional\end{tabular}}  \\ \hline

\multirow{3}{*}{\textbf{R18}} & \textbf{tiny} & 14 & 37 & 0.12 & 0.89 & 1.01 & 23.44 & 11.67 & 9.21  & \textbf{6.65} \\

 & \textbf{small} & 14 & 73 & 0.12 & 7.09 & 7.21 & 23.44 & 12.08 & 9.79  & 7.02 \\
 
 & \textbf{base} & 14 & 66 & 0.12 & 23.61 & 23.73 & 23.44 & 12.56 & 9.99  & 7.04 \\ \hline
 
\multirow{3}{*}{\textbf{R50}} & \textbf{tiny} & 21 & 37 & 0.14 & 0.89 & 1.03 & 19.62 & 11.67 & \textbf{8.79}  & 7.11 \\

 & \textbf{small} & 22 & 67 & 0.14 & 6.50 & 6.64 & 20.32 & 12.01 & 9.98  & 7.93 \\
 
 & \textbf{base} & 24 & 66 & 0.16 & 23.61 & 23.77 & 19.57 & 12.56 & 10.14  & 7.74 \\ \hline
 
\multirow{3}{*}{\textbf{R101}} & \textbf{tiny} & 19 & 56 & 0.10 & 1.33 & 1.43 & 20.04 & 11.48 & 9.26  & 7.13 \\

 & \textbf{small} & 22 & 67 & 0.14 & 6.50 & 6.64 & 19.42 & 12.01 & 9.88  & 7.82 \\
 
 & \textbf{base} & 18 & 66 & 0.10 & 23.61 & 23.71 & 20.87 & 12.56 & 10.07  & 7.30 \\ \hline
 
\end{tabular}%
}
\end{table}

\section{CONCLUSIONS AND FUTURE WORK}

We analyze the use of well-known pretrained Convolutional Neural Networks (CNNs) and Vision Transformers (ViTs) as out-of-the-box feature extraction methods for periocular recognition, examining the behavior of each network layer.
The models, pretrained on the ImageNet Large Scale Visual Recognition Challenge, are successful generic object recognizers proven to be very successful for various vision tasks.
While previous work focused on CNNs \cite{[Razavian14]}, this paper includes ViTs in the analysis.
We compare three CNNs based on ResNet (R18, R50 and R101) \cite{[He16]} and three ViTs (tiny, small, and base) \cite{dosovitskiy21iclr_ViT_transformers}, which share the same architecture but differ in depth or parameters.
Features from these pre-trained models are observed to be an effective way for periocular recognition, with the advantage of not needing to train the networks for the task.
Furthermore, CNNs and ViTs are observed to be highly complementary, with their fusion significantly improving performance.
We also demonstrate that traditional features encompassing Local Binary Patterns, Histogram of Oriented Gradients, and Scale-Invariant Feature Descriptors also complement the recognition task effectively.
%

We observe an upper limit in the number of CNN layers and parameters needed for optimal performance. The best layer of R18 provides an EER just 0.7\% worse than R50 but with 55\% fewer parameters. Engaging the much deeper R101 does not provide additional benefits and indeed has the worst performance among the three.
Regarding the depth needed to achieve optimum performance, the CNNs typically require nearly the entire network depth, while ViTs achieve close to optimal performance at just 30\% depth. 
%
%
In addition, the lighter versions, R18 CNN and tiny ViT, perform sufficiently well with significantly fewer parameters. Using larger pretrained models is thus not advantageous. 
This aligns with e.g. network pruning studies \cite{alonsofernandez2024deepnetworkpruningcomparative}, which show that large networks can be over-dimensionated, being possible to find lighter methods while not sacrificing performance.

When comparing feature vectors (using cosine similarity) we evaluate different strategies to L2-normalize the vectors.
Averaging feature vectors across channels or patches negatively impacts recognition accuracy. 
However, channel average is common in many CNN architectures (via Global Average Pooling layers, GAP) just before the fully connected classification section. In many transfer learning approaches, output from the GAP layer is often used as a feature vector as well. Our results question whether this strategy is optimal. 
Indeed, face recognition studies suggest alternatives such as Global Depth-wise Convolution (which performs a channel weighted average instead) or skipping averaging altogether \cite{Deng19CVPR_ArcFace}.

As future work, we plan to analyze the observed complementarity of the models by studying which parts of the ocular region they rely on \cite{Alonso23wifs_lime_biometrics}.
Extending this research to infra-red data (typical in iris recognition \cite{Nguyen24ACMCS_Iris_DL_survey}) or to other biometric modalities will verify if our conclusions hold across different domains.
Incorporating other CNN architectures that may be complementary to the present ones is another potential avenue for enhancing performance through network ensembles.

\section*{Acknowledgment}

Authors thank the Swedish Research Council (VR) and the Swedish Innovation Agency (VINNOVA) for funding their research. We gratefully acknowledge the support of NVIDIA Corporation with the donation of the Titan V GPU used for this research. The data handling in Sweden was enabled by the National Academic Infrastructure for Supercomputing in Sweden (NAISS).


{\small
\bibliographystyle{IEEEtran}


}

\end{document}